\documentclass[11pt]{article}

\usepackage[letterpaper,margin=1in]{geometry}
\usepackage[T1]{fontenc}
\usepackage[utf8]{inputenc}
\usepackage{times}
\usepackage{amsmath,amssymb}
\usepackage{booktabs,tabularx,array}
\usepackage{pgfplots}
\usepgfplotslibrary{groupplots}
\pgfplotsset{compat=1.12}
\usepackage{enumitem}
\usepackage{xcolor}
\usepackage{microtype}
\usepackage[round]{natbib}
\usepackage{url}
\usepackage[colorlinks=true,linkcolor=blue!55!black,citecolor=teal!60!black,urlcolor=teal!60!black]{hyperref}

\newcolumntype{Y}{>{\raggedright\arraybackslash}X}
\newcommand{\eC}{\textsc{ec}\textsuperscript{2}}

\title{\textbf{Decision-Focused Active Learning for\\Scale-Aware Critical-Materials Recovery}}
\author{Niranjan Srinivas\textsuperscript{1} \quad Debajyoti Ray\textsuperscript{1} \quad Elias Nakouzi\textsuperscript{2}\\[4pt]
{\normalsize \textsuperscript{1}Coactive Inc.\qquad \textsuperscript{2}Pacific Northwest National Laboratory}\\[2pt]
{\small Correspondence: \texttt{niranjan@coactive.science}}}
\date{Technical note --- September 2026}

\begin{document}
\maketitle

\begin{abstract}
Choosing a recovery process for scale-up requires connecting laboratory results with product requirements, process costs, and scale effects. We analyze records from Pacific Northwest National Laboratory's Computer Intelligence for Critical Element Recovery and Optimization (CICERO) workflow for autonomous selective precipitation. Active learning uses prior results to choose experiments. In a conditional retrospective benchmark with fitted models and recycled neodymium--iron--boron (NdFeB) magnet records, active learning finds the best recorded result with fewer experiments than nonadaptive space filling. Enrichment is the selected rare-earth-to-iron ratio relative to that in the feed. Adaptive policies reach the recorded enrichment maximum by 16--24 wells (individual experiments), versus 48. Our two-stage reconstruction ties two adaptive alternatives at 16 wells. Conditional analyses of recycled samarium--cobalt (SmCo) magnets show a Round~2 tradeoff between purity and nominal yield, the recovery fraction calculated from an assumed starting amount; NdFeB Round~1 routes differ in enrichment. Rankings for produced water from oil and gas extraction depend on phase and dilution assumptions requiring confirmation.

We propose choosing batches by their expected reduction in downstream Bayes risk: the minimum expected loss among available process decisions under current beliefs. In exploratory simulations, a hybrid that filters candidates has lower estimated loss than the implemented joint search across routes and conditions. Differences involving the synthetic two-stage policy are small relative to estimation uncertainty. We outline a pre-registered prospective test under a shared loss and logging standard, requiring clarified measurements and records, a defined process decision and relevant outputs, credible economic inputs, and validation at the intended scale.
\end{abstract}

\section{Background and motivation}

Choosing a recovery route and operating conditions for a larger scale requires more than laboratory purity or yield targets. The choice also depends on product requirements, process costs, and how laboratory results carry over to the intended scale. We ask how to select finite experimental batches to inform that choice.

Pacific Northwest National Laboratory (PNNL) developed Computer Intelligence for Critical Element Recovery and Optimization (CICERO), an agentic autonomous-laboratory workflow for selective precipitation from produced water (water from oil and gas extraction) and magnet leachates \citep{ritchhart2026materials}. It connects feedstock characterization, technoeconomic reasoning, planning, robotic execution, inductively coupled plasma mass spectrometry (ICP-MS), and Bayesian optimization. Its archive contains initial feed measurements, five per-well result files, liquid-handling scripts, experiment-agent records, qualitative technoeconomic analyses (TEA), and one saved Bayesian-optimization state.

In a companion preprint, two of this note's authors first distinguish recovery pathways using a policy motivated by \eC{} (equivalence-class edge cutting; \citealp{golovin2010near}). They then optimize continuous conditions within the selected route using batch Gaussian-process upper confidence bound (GP-UCB) optimization \citep{ray2026coactive,srinivas2010gaussian}. Their benchmark minimizes experimental cost to a laboratory target on CICERO-inspired synthetic surfaces, useful when resources constrain a campaign. The stages use different utilities, with no common experiment-selection criterion.

The PNNL co-author confirms that experimental cost was not material to route or condition selection in the CICERO campaigns considered here. Our focus is the eventual process choice, where product requirements and downstream process costs enter the decision.

Our NdFeB benchmark, using fitted models, supports faster search by adaptive policies than by the tested nonadaptive designs on the recorded pool. The SmCo tradeoff illustrates why selecting a condition to advance also requires a downstream criterion. We propose selecting batches by how much their results are expected to reduce the loss of the eventual process choice. Under current beliefs, the minimum expected downstream loss among available decisions is the downstream Bayes risk. A batch's value is the expected reduction in that risk after observing its results. This objective covers discrete routes and continuous conditions at the intended downstream scale.

Closing this experimental loop requires standardized, interoperable data with traceable provenance. Versioned pipelines should convert raw measurements and linked metadata into decision inputs before the next batch. We first review the records, then formulate the decision objective and benchmark search on the recorded NdFeB pool. Synthetic comparisons inform the proposed prospective test.

\section{What the current evidence supports}

SmCo illustrates a process-choice tradeoff, NdFeB supplies a conditional search target, and produced water identifies measurement and scale-transfer requirements.

We distinguish four evidence classes. \textsc{Observed} statements appear directly in the published paper or deposited records. \textsc{Derived} statements are calculations under stated assumptions. \textsc{Model-based} statements require a response, noise, economic, censoring, or scale-transfer model. \textsc{PNNL-confirmed} statements reflect the PNNL co-author's direct knowledge and interpretation of the experimental campaigns, limited by the deadline for submitting the current manuscript, which constrained the ability to collect more detailed information. Table~\ref{tab:evidence} summarizes the principal evidence sources; a claim-level evidence ledger linking each statement to source records, transformations, and assumptions is maintained by the authors and available on request.

\begin{table}[ht]
\centering
\caption{Principal evidence sources.}
\label{tab:evidence}
\small
\begin{tabularx}{\textwidth}{@{}p{0.16\textwidth}p{0.24\textwidth}Y Y@{}}
\toprule
Campaign & Record & Supported statement & Restriction \\
\midrule
SmCo Rounds 1--2 & CSVs, initial ICP-MS, protocols, and PNNL-author confirmation & \textsc{Derived}: purity--nominal-yield frontiers; \textsc{Model-based}: decision regions & Common feed and symbolic downstream loss \\
NdFeB Round 1 & Initial ICP-MS, CSV, protocol, and the companion preprint's two-stage method & \textsc{Derived}: route profiles; \textsc{Model-based}: finite-pool policies & Conditional map, signal semantics, and surrogate \\
Produced water & Initial ICP-MS, CSV, protocol, and agent JSON & \textsc{Derived}: row summaries; \textsc{Model-based}: interpretation sensitivities & Stock, map, units, dilution, and phase require confirmation \\
Produced water bench & Initial ICP-MS, protocol and agent records, and supplementary information & \textsc{Derived}: plate row B and bench run use similar NaOH:Mg ratios & Conditional on a common feed and recorded stocks \\
SmCo BO state & \path{smco_round2_bo_agent.state.json} & \textsc{Observed}: model and experiment-selection configuration & No recommendation list or selection history \\
TEA and UV--Vis & Agent JSON; supplementary information & \textsc{Observed}: qualitative economic factors and UV--Vis/ICP-MS slope 0.97 & No numerical decision rule, paired residuals, or relative assay cost \\
\bottomrule
\end{tabularx}
\end{table}

\subsection{SmCo: purity and nominal yield}

\begin{figure}[!hb]
\centering
\begin{tikzpicture}
\begin{groupplot}[
group style={group size=2 by 1, horizontal sep=0.8cm},
width=0.43\textwidth, height=0.32\textwidth,
xmin=0, xmax=102, ymin=0, ymax=120,
xtick={0,20,40,60,80,100}, ytick={0,20,40,60,80,100,120},
xlabel={Binary Sm--Co purity (\%)},
tick label style={font=\scriptsize}, label style={font=\small},
title style={font=\small}, grid=major, grid style={gray!20},
clip=false,
legend style={at={(1.08,-0.42)}, anchor=north, legend columns=4, draw=none, font=\scriptsize}
]
\nextgroupplot[title={(a) Round 1}, ylabel={Raw nominal Sm yield (\%)}]
\addplot[forget plot, only marks, mark=*, mark size=1.1pt, gray!55] coordinates {(80.3483,21.5567) (67.725,24.8003) (75.3952,40.1745) (84.6729,65.9011) (64.171,60.2482) (49.4489,73.4873) (34.9666,72.8481) (29.7556,69.7688) (25.7433,63.4913) (25.5235,73.6197) (22.3993,84.6209) (20.3107,68.3962) (68.8172,44.0213) (75.8908,59.634) (74.8463,87.3465) (64.1067,67.7787) (50.5181,80.0241) (40.8441,65.5742) (35.4099,79.0726) (29.4198,68.0742) (24.3362,65.7303) (22.6086,71.5342) (22.1586,66.821) (20.2506,68.4475) (35.7584,78.7911) (41.4643,98.1022) (40.0944,78.5343) (37.7528,92.8797) (31.107,85.0463) (28.7081,85.2892) (25.7737,75.2939) (24.8758,69.5431) (23.1122,78.816) (21.734,70.3527) (20.9112,64.7656) (20.7472,63.5979) (22.2018,66.8339) (22.8011,89.6402) (22.1574,82.1527) (21.5204,80.9581) (20.7555,74.4059) (20.5415,82.5369) (20.3872,61.8499) (20.5132,67.5603) (20.5819,74.9603) (20.2993,62.5125) (20.368,66.5925) (20.2753,71.9634) (19.6101,74.1199) (19.8904,83.3927) (19.5394,82.0567) (19.6549,60.4665) (19.8744,63.8) (19.6489,94.0713) (19.8584,84.5096) (19.789,74.8988) (20.1127,78.439) (19.9734,85.9081) (20.2192,87.0159) (19.8792,66.2674) (19.7809,96.0234) (19.8015,77.6259) (19.8123,66.047) (19.8682,87.1202) (19.5209,72.9882) (19.7923,75.2371) (19.6434,88.78) (19.6399,76.581) (19.9521,82.9864) (19.7613,75.8656) (19.7599,65.7266) (19.9162,77.1568) (20.2663,77.9105) (20.0473,79.1386) (19.8107,77.0345) (19.5935,79.2175) (19.6092,81.7826) (19.6972,76.6883) (19.6173,78.7171) (19.6996,72.1275) (19.3132,67.3347) (19.7671,35.9981) (19.3173,87.8867) (19.4586,76.9876) (19.9975,84.6185) (20.7365,87.0764) (20.7693,77.6196) (21.2156,75.2003) (20.6482,67.0417) (20.5018,18.7598) (19.7683,69.5913) (19.7744,78.2396) (19.6589,82.5768) (19.1927,78.4479) (19.017,84.7554) (18.8515,106.741)};
\addplot[forget plot, blue!70!black, line width=0.8pt, no marks] coordinates {(18.8515,106.741) (41.4643,98.1022) (74.8463,87.3465) (84.6729,65.9011)};
\addplot[forget plot, only marks, mark=*, mark size=2.1pt, blue!70!black] coordinates {(84.6729,65.9011) (74.8463,87.3465) (41.4643,98.1022)};
\addplot[forget plot, only marks, mark=triangle*, mark size=2.5pt, orange!85!black] coordinates {(18.8515,106.741)};
\addplot[forget plot, red!65!black, dashed, line width=0.6pt, no marks] coordinates {(0,100) (102,100)};
\node[font=\fontsize{6}{7}\selectfont, anchor=south west] at (axis cs:84.6729,65.9011) {A04};
\node[font=\fontsize{6}{7}\selectfont, anchor=south west] at (axis cs:74.8463,87.3465) {B03};
\node[font=\fontsize{6}{7}\selectfont, anchor=south west] at (axis cs:41.4643,98.1022) {C02};
\node[font=\fontsize{6}{7}\selectfont, anchor=south west] at (axis cs:18.8515,106.741) {H12};
\addlegendimage{only marks, mark=*, gray!55}
\addlegendentry{Recorded condition}
\addlegendimage{blue!70!black, line width=0.8pt, mark=*}
\addlegendentry{Pareto frontier}
\addlegendimage{only marks, mark=triangle*, orange!85!black}
\addlegendentry{$Y_{\mathrm{nom}}>100\%$}
\addlegendimage{red!65!black, dashed, line width=0.6pt}
\addlegendentry{$100\%$ nominal yield}
\nextgroupplot[title={(b) Round 2}]
\addplot[forget plot, only marks, mark=*, mark size=1.1pt, gray!55] coordinates {(28.0952,0.00366277) (40.5877,0.00654171) (28.0952,0.00366277) (53.163,0.0106403) (55.3138,0.0116037) (28.0952,0.00366277) (28.0952,0.00366277) (28.0952,0.00366277) (68.9264,0.0207935) (77.213,0.0901151) (95.575,29.4903) (94.2781,95.5646) (85.0823,0.0534654) (71.3187,0.0233099) (74.4498,0.0273151) (77.8112,0.0328734) (28.0952,0.00366277) (59.564,0.0138086) (35.3024,0.00511506) (28.0952,0.00366277) (55.643,0.0117593) (81.0266,0.0876922) (90.3093,30.6385) (96.9191,84.3603) (77.1683,0.0667576) (69.369,0.0212294) (23.5622,0.00672851) (31.17,0.00424515) (29.0717,0.00384225) (75.7819,0.0293333) (65.6781,0.0179384) (40.0532,0.00626334) (41.1439,0.00728708) (89.8619,19.7024) (93.8859,80.7757) (87.2069,89.6519) (52.5074,0.0770464) (38.8275,0.0319723) (26.2567,0.0168506) (8.94186,0.00616994) (12.9712,0.00489712) (28.0952,0.00366277) (6.31239,0.014087) (12.2399,0.00458579) (41.061,0.168852) (82.5457,87.9507) (93.1973,68.8891) (49.2461,100.654) (19.8289,0.014739) (9.39028,0.00915143) (12.812,0.0176893) (6.26546,0.00539343) (19.6276,0.0199566) (22.2337,0.0293956) (8.00799,0.00517732) (28.1622,0.117952) (93.7529,20.6261) (92.2179,96.3101) (70.6667,103.598) (58.1018,103.299) (30.7822,0.0338678) (17.5295,0.0278755) (8.17239,0.0127519) (6.80067,0.0175025) (4.34805,0.00849946) (4.61959,0.011882) (13.852,0.0383382) (48.5144,0.459535) (29.3945,0.00390268) (48.9357,95.3165) (85.1429,98.7757) (37.1206,107.833) (18.0741,0.0180007) (10.8383,0.0300475) (6.92022,0.0205463) (4.21058,0.00672851) (3.99034,0.0174403) (7.6613,0.0369116) (4.31257,0.0243647) (94.0931,50.3751) (94.4131,92.9424) (66.4964,97.4717) (47.0081,96.042) (44.8181,57.9051) (11.7062,0.00446308) (9.78326,0.00505279) (4.58069,0.00669737) (3.99074,0.0263848) (3.6724,0.0165722) (3.11124,0.015206) (84.9185,11.0573) (91.1158,90.1227) (37.2112,29.3712) (55.8654,106.08) (41.8233,113.865) (38.0943,46.9821)};
\addplot[forget plot, blue!70!black, line width=0.8pt, no marks] coordinates {(41.8233,113.865) (55.8654,106.08) (70.6667,103.598) (85.1429,98.7757) (92.2179,96.3101) (94.2781,95.5646) (94.4131,92.9424) (96.9191,84.3603)};
\addplot[forget plot, only marks, mark=*, mark size=2.1pt, blue!70!black] coordinates {(96.9191,84.3603) (94.4131,92.9424) (94.2781,95.5646) (92.2179,96.3101) (85.1429,98.7757)};
\addplot[forget plot, only marks, mark=triangle*, mark size=2.5pt, orange!85!black] coordinates {(70.6667,103.598) (55.8654,106.08) (41.8233,113.865)};
\addplot[forget plot, red!65!black, dashed, line width=0.6pt, no marks] coordinates {(0,100) (102,100)};
\draw[gray!60, line width=0.3pt] (axis cs:96.9191,84.3603) -- (axis cs:91,75);
\node[font=\fontsize{6}{7}\selectfont, anchor=center, fill=white, inner sep=0.4pt] at (axis cs:91,75) {B12};
\draw[gray!60, line width=0.3pt] (axis cs:94.4131,92.9424) -- (axis cs:84,91);
\node[font=\fontsize{6}{7}\selectfont, anchor=east, fill=white, inner sep=0.4pt] at (axis cs:84,91) {G09};
\draw[gray!60, line width=0.3pt] (axis cs:94.2781,95.5646) -- (axis cs:101,88);
\node[font=\fontsize{6}{7}\selectfont, anchor=east, fill=white, inner sep=0.4pt] at (axis cs:101,88) {A12};
\draw[gray!60, line width=0.3pt] (axis cs:92.2179,96.3101) -- (axis cs:100,104);
\node[font=\fontsize{6}{7}\selectfont, anchor=east, fill=white, inner sep=0.4pt] at (axis cs:100,104) {E10};
\draw[gray!60, line width=0.3pt] (axis cs:85.1429,98.7757) -- (axis cs:86,108);
\node[font=\fontsize{6}{7}\selectfont, anchor=center, fill=white, inner sep=0.4pt] at (axis cs:86,108) {F11};
\draw[gray!60, line width=0.3pt] (axis cs:70.6667,103.598) -- (axis cs:72,112);
\node[font=\fontsize{6}{7}\selectfont, anchor=center, fill=white, inner sep=0.4pt] at (axis cs:72,112) {E11};
\draw[gray!60, line width=0.3pt] (axis cs:55.8654,106.08) -- (axis cs:58,115);
\node[font=\fontsize{6}{7}\selectfont, anchor=center, fill=white, inner sep=0.4pt] at (axis cs:58,115) {H10};
\draw[gray!60, line width=0.3pt] (axis cs:41.8233,113.865) -- (axis cs:42,118);
\node[font=\fontsize{6}{7}\selectfont, anchor=center, fill=white, inner sep=0.4pt] at (axis cs:42,118) {H11};
\end{groupplot}
\end{tikzpicture}
\caption{Derived recorded SmCo purity--raw-nominal-yield frontiers. Gray points are recorded conditions; labels identify nondominated wells. Values above 100\% remain unmodified.}
\label{fig:smco-frontiers}
\end{figure}
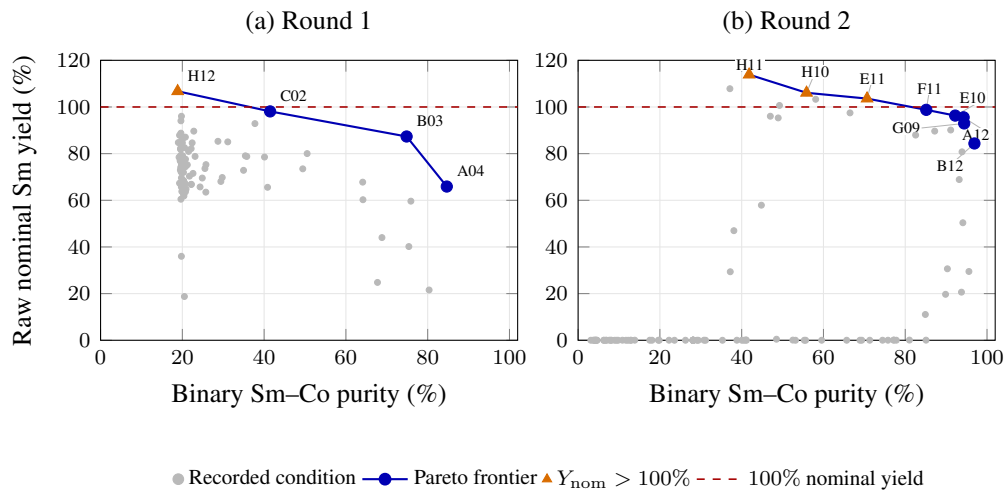

\textsc{PNNL-confirmed}. The paper's purity metric is $n_{\mathrm{Sm}}/(n_{\mathrm{Sm}}+n_{\mathrm{Co}})$, and the reported moles include the ICP-MS dilution corrections (20{,}000$\times$ in Round~1 and 34{,}000$\times$ in Round~2). \textsc{Observed}. Both protocols use 200~$\mu$L of leachate per well. \textsc{Derived}. Assuming the initial Sm concentration of 3928.381~mg/L applies to both rounds gives 5.225~$\mu$mol nominal starting Sm per well. We define
\[
Y_{\mathrm{nom}}=100\,\frac{n_{\mathrm{Sm,recovered}}}{5.2253006~\mu\mathrm{mol}}.
\]
\textsc{Derived}. The recorded purity--yield frontiers contain four Round~1 and eight Round~2 conditions (Fig.~\ref{fig:smco-frontiers}). Among the five Round~2 frontier points at or below 100\% nominal yield, purity decreases from 96.92\% to 85.14\% as nominal yield increases from 84.36\% to 98.78\%; three additional frontier points exceed 100\% and are reported uncorrected pending confirmation of the feed concentration.

The frontier characterizes the purity--yield tradeoff and provides a target for further chemistry development. Selecting one condition to advance requires an explicit downstream criterion.

\textsc{PNNL-confirmed}. Purity is binary Sm--Co purity, not total elemental purity, and Nd and Dy were not tracked during purification. \textsc{Derived}. Initial Nd and Dy together account for about 0.2~mol\%; the raw calculation gives one Round~1 and six Round~2 wells above 100\% nominal yield.

\begin{samepage}
\textsc{Model-based}. To express the downstream choice symbolically, let
\[
\ell_i(g,\rho)=1-r_i+\rho r_i\mathbf{1}(p_i<g),
\]
where $r_i$ is the raw nominal-yield fraction, $p_i$ is binary Sm--Co purity, $g\in[0,1]$ is the required grade, and $\rho\geq0$ is the below-grade consequence per recovered Sm normalized by the value of recovered Sm; not deploying has loss one.
\end{samepage}

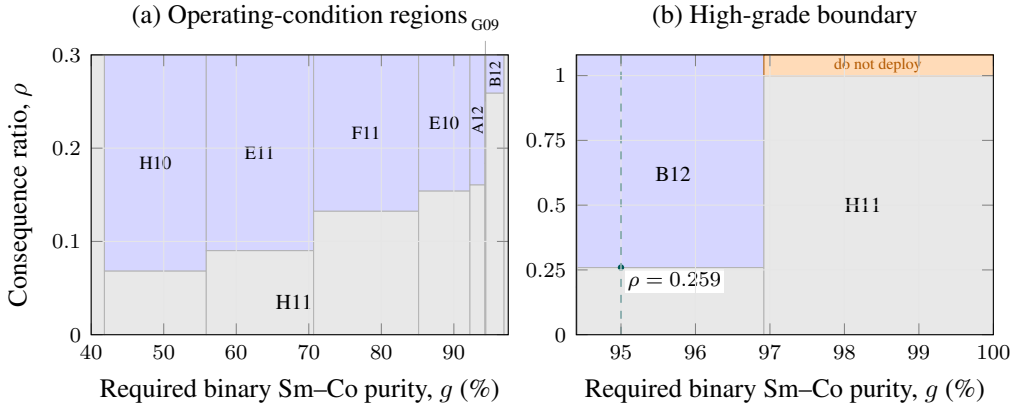
\begin{figure}[!htbp]
\centering
\begin{tikzpicture}
\begin{groupplot}[
group style={group size=2 by 1, horizontal sep=0.9cm},
width=0.43\textwidth, height=0.32\textwidth,
xlabel={Required binary Sm--Co purity, $g$ (\%)},
tick label style={font=\scriptsize}, label style={font=\small},
title style={font=\small}, grid=major, grid style={gray!20},
axis on top, clip=false
]
\nextgroupplot[title={(a) Operating-condition regions},
xmin=40, xmax=97.5, ymin=0, ymax=0.3,
xtick={40,50,60,70,80,90}, ytick={0,0.1,0.2,0.3},
ylabel={Consequence ratio, $\rho$}]
\path[fill=gray!18, draw=gray!60, line width=0.25pt] (axis cs:40,0) rectangle (axis cs:41.8232727,0.3);
\path[fill=gray!18, draw=gray!60, line width=0.25pt] (axis cs:41.8232727,0) rectangle (axis cs:55.8654318,0.0683710253);
\path[fill=blue!16, draw=gray!60, line width=0.25pt] (axis cs:41.8232727,0.0683710253) rectangle (axis cs:55.8654318,0.3);
\path[fill=gray!18, draw=gray!60, line width=0.25pt] (axis cs:55.8654318,0) rectangle (axis cs:70.6666999,0.0901676941);
\path[fill=blue!16, draw=gray!60, line width=0.25pt] (axis cs:55.8654318,0.0901676941) rectangle (axis cs:70.6666999,0.3);
\path[fill=gray!18, draw=gray!60, line width=0.25pt] (axis cs:70.6666999,0) rectangle (axis cs:85.1428776,0.132516477);
\path[fill=blue!16, draw=gray!60, line width=0.25pt] (axis cs:70.6666999,0.132516477) rectangle (axis cs:85.1428776,0.3);
\path[fill=gray!18, draw=gray!60, line width=0.25pt] (axis cs:85.1428776,0) rectangle (axis cs:92.2179062,0.154170224);
\path[fill=blue!16, draw=gray!60, line width=0.25pt] (axis cs:85.1428776,0.154170224) rectangle (axis cs:92.2179062,0.3);
\path[fill=gray!18, draw=gray!60, line width=0.25pt] (axis cs:92.2179062,0) rectangle (axis cs:94.2780749,0.160716815);
\path[fill=blue!16, draw=gray!60, line width=0.25pt] (axis cs:92.2179062,0.160716815) rectangle (axis cs:94.2780749,0.3);
\path[fill=gray!18, draw=gray!60, line width=0.25pt] (axis cs:94.2780749,0) rectangle (axis cs:94.4130621,0.183746667);
\path[fill=blue!16, draw=gray!60, line width=0.25pt] (axis cs:94.2780749,0.183746667) rectangle (axis cs:94.4130621,0.3);
\path[fill=gray!18, draw=gray!60, line width=0.25pt] (axis cs:94.4130621,0) rectangle (axis cs:96.9190869,0.259117173);
\path[fill=blue!16, draw=gray!60, line width=0.25pt] (axis cs:94.4130621,0.259117173) rectangle (axis cs:96.9190869,0.3);
\path[fill=gray!18, draw=gray!60, line width=0.25pt] (axis cs:96.9190869,0) rectangle (axis cs:97.5,0.3);
\node[font=\scriptsize] at (axis cs:68,0.035) {H11};
\node[font=\fontsize{7}{8}\selectfont] at (axis cs:48.8443523,0.184185513) {H10};
\node[font=\fontsize{7}{8}\selectfont] at (axis cs:63.2660659,0.195083847) {E11};
\node[font=\fontsize{7}{8}\selectfont] at (axis cs:77.9047887,0.216258238) {F11};
\node[font=\fontsize{7}{8}\selectfont] at (axis cs:88.6803919,0.227085112) {E10};
\node[font=\fontsize{6}{7}\selectfont, rotate=90] at (axis cs:93.2479905,0.230358407) {A12};
\draw[gray!65, line width=0.3pt] (axis cs:94.3455685,0.25) -- (axis cs:94.3455685,0.315);
\node[font=\fontsize{6}{7}\selectfont, anchor=south] at (axis cs:94.3455685,0.315) {G09};
\node[font=\fontsize{6}{7}\selectfont, rotate=90] at (axis cs:95.6660745,0.279558587) {B12};
\nextgroupplot[title={(b) High-grade boundary},
xmin=94.4, xmax=100, ymin=0, ymax=1.08,
xtick={95,96,97,98,99,100}, ytick={0,0.25,0.5,0.75,1},]
\path[fill=gray!18, draw=gray!60, line width=0.25pt] (axis cs:94.4,0) rectangle (axis cs:94.4130621,0.183746667);
\path[fill=blue!16, draw=gray!60, line width=0.25pt] (axis cs:94.4,0.183746667) rectangle (axis cs:94.4130621,1.08);
\path[fill=gray!18, draw=gray!60, line width=0.25pt] (axis cs:94.4130621,0) rectangle (axis cs:96.9190869,0.259117173);
\path[fill=blue!16, draw=gray!60, line width=0.25pt] (axis cs:94.4130621,0.259117173) rectangle (axis cs:96.9190869,1.08);
\path[fill=gray!18, draw=gray!60, line width=0.25pt] (axis cs:96.9190869,0) rectangle (axis cs:100,1);
\path[fill=orange!28, draw=orange!70!black, line width=0.25pt] (axis cs:96.9190869,1) rectangle (axis cs:100,1.08);
\draw[teal!70!black, dashed, line width=0.6pt] (axis cs:95,0) -- (axis cs:95,1.08);
\fill[teal!70!black] (axis cs:95,0.259117173) circle[radius=1.2pt];
\node[font=\scriptsize, anchor=north west, fill=white, inner sep=1pt] at (axis cs:95.05,0.252117173) {$\rho=0.259$};
\node[font=\scriptsize] at (axis cs:95.7,0.62) {B12};
\node[font=\scriptsize] at (axis cs:98.25,0.5) {H11};
\node[font=\fontsize{6}{7}\selectfont, text=orange!70!black] at (axis cs:98.45,1.04) {do not deploy};
\end{groupplot}
\end{tikzpicture}
\caption{Model-based Round~2 decision regions using raw nominal yield. Panel (a) resolves the operating-condition boundaries over the range where recorded optima switch; panel (b) shows the 95\%-grade break-even and the high-grade do-not-deploy region.}
\label{fig:smco-decision-regions}
\end{figure}

When the below-grade consequence is small, greater nominal recovery can outweigh lower purity in this model. For Round~2 at $g=0.95$, the loss-minimizing recorded condition switches from H11 to B12 at $\rho=0.259$ (tie at the boundary): H11 has 113.86\% nominal yield and 41.82\% purity, whereas B12 has 84.36\% nominal yield and 96.92\% purity. At this boundary, the consequence per recovered Sm is about a quarter of the value assigned to that Sm. If $g>0.9692$, no recorded condition meets grade and not deploying is optimal for $\rho>1$ (Fig.~\ref{fig:smco-decision-regions}). Exact intervals and boundary ties are retained in machine-readable form and are available from the authors.

\subsection{NdFeB Round 1: conditional route contrast}
\label{sec:ndfeb-records}

\textsc{Observed}. The deposited \path{ndfeb_round1.py} assigns 1~M NaOH to columns 1--6 and 0.2~M sodium oxalate to columns 7--12, giving 48 wells per route. Reagent volume increases from 30 to 180~$\mu$L within each block, and leachate volume increases by row. The experiment-agent JSON describes a different map and oxalate concentration, so the script must be confirmed before the mapping is treated as executed.

Conditional on that script and on mass-proportional signals, define atomic-weight-adjusted signal proxies $\widetilde n_j^S=z_j^S/M_j$ and normalized rare-earth-element (REE)/Fe enrichment
\[
\widetilde N_R^S=\widetilde n_{\mathrm{Pr}}^S+\widetilde n_{\mathrm{Nd}}^S+\widetilde n_{\mathrm{Gd}}^S+\widetilde n_{\mathrm{Tb}}^S+\widetilde n_{\mathrm{Dy}}^S,
\qquad
E=\frac{\widetilde N_R^{P}/\widetilde n_{\mathrm{Fe}}^{P}}{\widetilde N_R^{F}/\widetilde n_{\mathrm{Fe}}^{F}},
\]
where $z_j^S$ is the recorded signal for element $j$ in sample $S$, $M_j$ is its atomic weight, and $P$ and $F$ denote the recorded Round~1 plate sample and initial feed. The calculation uses elements present in both files and assumes common element-wise dilution; the plate phase and signal units remain unconfirmed.

We use the NdFeB enrichment score as a reproducible benchmark target because it can be calculated from the recorded measurements under these assumptions.

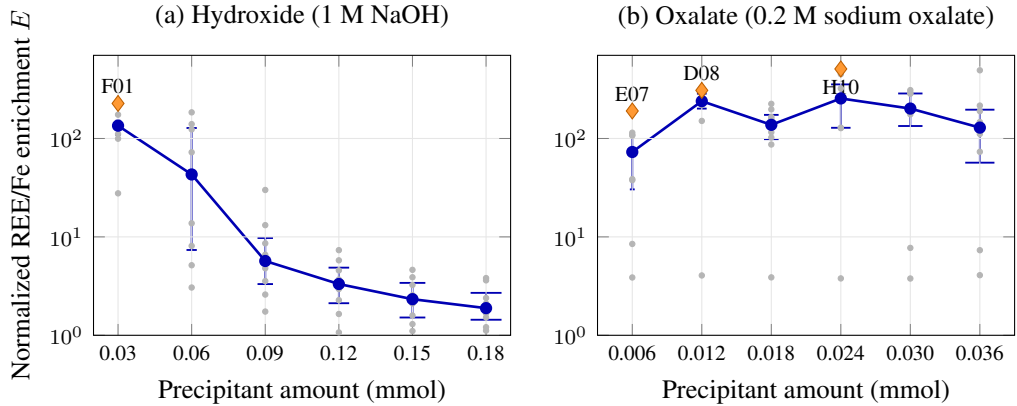
\begin{figure}[ht]
\centering
\begin{tikzpicture}
\begin{groupplot}[
group style={group size=2 by 1, horizontal sep=1.15cm},
width=0.43\textwidth, height=0.32\textwidth,
ymode=log, log basis y=10, ymin=1, ymax=700, ytick={1,10,100},
scaled x ticks=false,
xlabel={Precipitant amount (mmol)},
tick label style={font=\scriptsize}, label style={font=\small},
title style={font=\small}, grid=major, grid style={gray!20}, axis on top
]
\nextgroupplot[title={(a) Hydroxide (1 M NaOH)},xmin=0.02, xmax=0.19,xtick={0.03,0.06,0.09,0.12,0.15,0.18},xticklabels={0.03,0.06,0.09,0.12,0.15,0.18},ylabel={Normalized REE/Fe enrichment $E$}]
\addplot[only marks, mark=*, mark size=1pt, gray!58] coordinates {(0.03,27.7102159) (0.06,3.05849699) (0.09,1.74269822) (0.12,1.0695929) (0.15,1.10495573) (0.18,1.11614811) (0.03,143.992585) (0.06,5.15052557) (0.09,2.59453444) (0.12,1.64641709) (0.15,1.29931169) (0.18,1.20864255) (0.03,99.5014236) (0.06,8.09927177) (0.09,3.55567336) (0.12,2.2757358) (0.15,1.58873928) (0.18,1.51561882) (0.03,110.995307) (0.06,13.7532021) (0.09,4.79717827) (0.12,2.95296217) (0.15,2.10188922) (0.18,1.67136853) (0.03,174.630378) (0.06,72.4423783) (0.09,6.58328845) (0.12,3.68549699) (0.15,2.55492681) (0.18,2.10376598) (0.03,226.984476) (0.06,123.769979) (0.09,8.59485085) (0.12,4.56559514) (0.15,3.24838615) (0.18,2.39276884) (0.03,143.162583) (0.06,140.007895) (0.09,13.1588524) (0.12,5.77740564) (0.15,3.88895268) (0.18,3.81488975) (0.03,126.673251) (0.06,184.321136) (0.09,29.9774244) (0.12,7.33205581) (0.15,4.62464835) (0.18,3.61655632)};
\draw[blue!70!black, line width=0.7pt] (axis cs:0.03,108.121836) -- (axis cs:0.03,151.652034);
\draw[blue!70!black, line width=0.7pt] (axis cs:0.02895,108.121836) -- (axis cs:0.03105,108.121836);
\draw[blue!70!black, line width=0.7pt] (axis cs:0.02895,151.652034) -- (axis cs:0.03105,151.652034);
\draw[blue!70!black, line width=0.7pt] (axis cs:0.06,7.36208522) -- (axis cs:0.06,127.829458);
\draw[blue!70!black, line width=0.7pt] (axis cs:0.0579,7.36208522) -- (axis cs:0.0621,7.36208522);
\draw[blue!70!black, line width=0.7pt] (axis cs:0.0579,127.829458) -- (axis cs:0.0621,127.829458);
\draw[blue!70!black, line width=0.7pt] (axis cs:0.09,3.31538863) -- (axis cs:0.09,9.73585124);
\draw[blue!70!black, line width=0.7pt] (axis cs:0.08685,3.31538863) -- (axis cs:0.09315,3.31538863);
\draw[blue!70!black, line width=0.7pt] (axis cs:0.08685,9.73585124) -- (axis cs:0.09315,9.73585124);
\draw[blue!70!black, line width=0.7pt] (axis cs:0.12,2.11840612) -- (axis cs:0.12,4.86854777);
\draw[blue!70!black, line width=0.7pt] (axis cs:0.1158,2.11840612) -- (axis cs:0.1242,2.11840612);
\draw[blue!70!black, line width=0.7pt] (axis cs:0.1158,4.86854777) -- (axis cs:0.1242,4.86854777);
\draw[blue!70!black, line width=0.7pt] (axis cs:0.15,1.51638238) -- (axis cs:0.15,3.40852779);
\draw[blue!70!black, line width=0.7pt] (axis cs:0.14475,1.51638238) -- (axis cs:0.15525,1.51638238);
\draw[blue!70!black, line width=0.7pt] (axis cs:0.14475,3.40852779) -- (axis cs:0.15525,3.40852779);
\draw[blue!70!black, line width=0.7pt] (axis cs:0.18,1.43887475) -- (axis cs:0.18,2.69871571);
\draw[blue!70!black, line width=0.7pt] (axis cs:0.1737,1.43887475) -- (axis cs:0.1863,1.43887475);
\draw[blue!70!black, line width=0.7pt] (axis cs:0.1737,2.69871571) -- (axis cs:0.1863,2.69871571);
\addplot[blue!70!black, line width=1pt, mark=*, mark size=1.8pt] coordinates {(0.03,134.917917) (0.06,43.0977902) (0.09,5.69023336) (0.12,3.31922958) (0.15,2.32840802) (0.18,1.88756725)};
\addplot[only marks, mark=diamond*, mark size=3pt, mark options={fill=orange!80, draw=orange!75!black}] coordinates {(0.03,226.984476)};
\node[font=\scriptsize, anchor=south] at (axis cs:0.03,226.984476) {F01};
\nextgroupplot[title={(b) Oxalate (0.2 M sodium oxalate)},xmin=0.003, xmax=0.039,xtick={0.006,0.012,0.018,0.024,0.030,0.036},xticklabels={0.006,0.012,0.018,0.024,0.030,0.036}]
\addplot[only marks, mark=*, mark size=1pt, gray!58] coordinates {(0.006,108.639591) (0.012,150.831736) (0.018,115.697915) (0.024,450.27696) (0.03,7.73087791) (0.036,4.08103824) (0.006,8.47891746) (0.012,218.12892) (0.018,86.9018865) (0.024,128.748634) (0.03,207.965288) (0.036,7.32242793) (0.006,38.6339271) (0.012,256.274046) (0.018,102.296004) (0.024,127.474486) (0.03,284.8397) (0.036,73.2676902) (0.006,37.6944981) (0.012,308.435857) (0.018,159.612584) (0.024,267.504978) (0.03,175.982177) (0.036,146.849614) (0.006,190.438713) (0.012,286.27516) (0.018,165.83802) (0.024,322.714061) (0.03,292.228272) (0.036,189.572502) (0.006,107.167538) (0.012,222.911857) (0.018,225.000078) (0.024,242.93572) (0.03,310.115694) (0.036,111.10955) (0.006,3.86865459) (0.012,4.05636109) (0.018,3.87807176) (0.024,3.78527831) (0.03,3.77612442) (0.036,216.009363) (0.006,114.504707) (0.012,285.272852) (0.018,197.609893) (0.024,508.966758) (0.03,194.928599) (0.036,493.110987)};
\draw[blue!70!black, line width=0.7pt] (axis cs:0.006,30.390603) -- (axis cs:0.006,110.10587);
\draw[blue!70!black, line width=0.7pt] (axis cs:0.00579,30.390603) -- (axis cs:0.00621,30.390603);
\draw[blue!70!black, line width=0.7pt] (axis cs:0.00579,110.10587) -- (axis cs:0.00621,110.10587);
\draw[blue!70!black, line width=0.7pt] (axis cs:0.012,201.304624) -- (axis cs:0.012,285.523429);
\draw[blue!70!black, line width=0.7pt] (axis cs:0.01158,201.304624) -- (axis cs:0.01242,201.304624);
\draw[blue!70!black, line width=0.7pt] (axis cs:0.01158,285.523429) -- (axis cs:0.01242,285.523429);
\draw[blue!70!black, line width=0.7pt] (axis cs:0.018,98.4474744) -- (axis cs:0.018,173.780988);
\draw[blue!70!black, line width=0.7pt] (axis cs:0.01737,98.4474744) -- (axis cs:0.01863,98.4474744);
\draw[blue!70!black, line width=0.7pt] (axis cs:0.01737,173.780988) -- (axis cs:0.01863,173.780988);
\draw[blue!70!black, line width=0.7pt] (axis cs:0.024,128.430097) -- (axis cs:0.024,354.604786);
\draw[blue!70!black, line width=0.7pt] (axis cs:0.02316,128.430097) -- (axis cs:0.02484,128.430097);
\draw[blue!70!black, line width=0.7pt] (axis cs:0.02316,354.604786) -- (axis cs:0.02484,354.604786);
\draw[blue!70!black, line width=0.7pt] (axis cs:0.03,133.919352) -- (axis cs:0.03,286.686843);
\draw[blue!70!black, line width=0.7pt] (axis cs:0.02895,133.919352) -- (axis cs:0.03105,133.919352);
\draw[blue!70!black, line width=0.7pt] (axis cs:0.02895,286.686843) -- (axis cs:0.03105,286.686843);
\draw[blue!70!black, line width=0.7pt] (axis cs:0.036,56.7813747) -- (axis cs:0.036,196.181717);
\draw[blue!70!black, line width=0.7pt] (axis cs:0.03474,56.7813747) -- (axis cs:0.03726,56.7813747);
\draw[blue!70!black, line width=0.7pt] (axis cs:0.03474,196.181717) -- (axis cs:0.03726,196.181717);
\addplot[blue!70!black, line width=1pt, mark=*, mark size=1.8pt] coordinates {(0.006,72.9007325) (0.012,239.592952) (0.018,137.655249) (0.024,255.220349) (0.03,201.446944) (0.036,128.979582)};
\addplot[only marks, mark=diamond*, mark size=3pt, mark options={fill=orange!80, draw=orange!75!black}] coordinates {(0.012,308.435857) (0.006,190.438713) (0.024,508.966758)};
\node[font=\scriptsize, anchor=south] at (axis cs:0.012,308.435857) {D08};
\node[font=\scriptsize, anchor=south] at (axis cs:0.006,190.438713) {E07};
\node[font=\scriptsize, anchor=north] at (axis cs:0.024,508.966758) {H10};
\end{groupplot}
\end{tikzpicture}
\caption{Derived conditional NdFeB Round~1 profiles. Gray points are enrichment values for individual recorded wells, blue lines and bars are medians and interquartile ranges, and labeled diamonds are within-route frontiers. The deposited-script map and concentrations remain unconfirmed.}
\label{fig:ndfeb-route-profiles}
\end{figure}

\textsc{Derived}. Across increasing reagent-volume columns, median $E$ values are approximately 135, 43, 5.7, 3.3, 2.3, and 1.9 for hydroxide, and 73, 240, 138, 255, 201, and 129 for oxalate. Maximizing $E$ while minimizing precipitant amount within each route leaves F01 on the hydroxide frontier and E07, D08, and H10 on the oxalate frontier (Fig.~\ref{fig:ndfeb-route-profiles}). \textsc{Model-based}. Quadratic ridge fits have nested leave-one-out root-mean-square errors in log enrichment (log-RMSE) of 0.536 for hydroxide and 1.606 for oxalate; the latter provides little support for smooth interpolation of the oxalate response. These results are conditional normalized-enrichment contrasts, not recovery, a confirmed separation factor, or economic preference.

\subsection{Produced water: replication and plate-to-bench correspondence}

\textsc{Observed}. The deposited script assigns 30--210~$\mu$L NaOH to rows A--G and no NaOH to row H; each row has 12 wells. The experiment-agent record specifies 1~M NaOH. The supplementary information reports a 20~mL bench experiment using 9~mL of 0.5~M NaOH and a product containing 98.68~mol\% Mg \citep{ritchhart2026materials}. \textsc{Derived}. Under the recorded stock concentrations and deposited map, the shared initial Mg concentration of 1578.838~mg/L gives 3.42~mol NaOH per mol Mg in plate row B and, under a common-feed assumption, 3.46 in the bench experiment. Equivalently, the NaOH doses are 1.71 and 1.73 times the formal 2:1 OH:Mg stoichiometry of Mg(OH)$_2$. These dose ratios alone determine neither aqueous hydroxide availability nor precipitation yield.

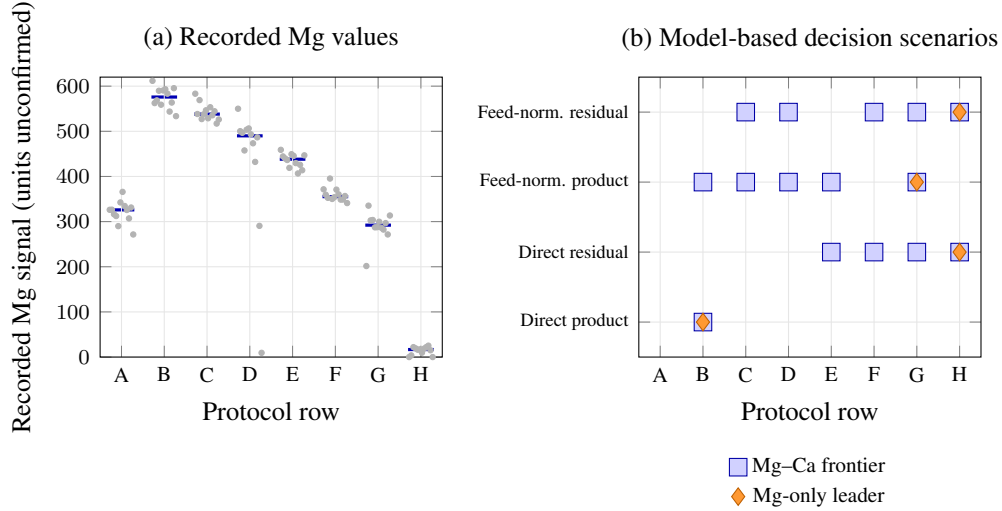
\begin{figure}[ht]
\centering
\begin{tikzpicture}
\begin{groupplot}[
group style={group size=2 by 1, horizontal sep=2.6cm},
width=0.37\textwidth, height=0.32\textwidth,
xmin=0.5, xmax=8.5, xtick={1,2,3,4,5,6,7,8},
xticklabels={A,B,C,D,E,F,G,H},
xlabel={Protocol row},
tick label style={font=\scriptsize}, label style={font=\small},
title style={font=\small}, axis on top
]
\nextgroupplot[title={(a) Recorded Mg values},
ymin=0, ymax=620, ytick={0,100,200,300,400,500,600},
ylabel={Recorded Mg signal (units unconfirmed)},
grid=major, grid style={gray!20}]
\addplot[only marks, mark=*, mark size=1pt, gray!60] coordinates {(0.725,326.16629) (0.775,326.3362) (0.825,316.2767) (0.875,312.38069) (0.925,289.99907) (0.975,342.46494) (1.025,365.84974) (1.075,334.52259) (1.125,325.7571) (1.175,307.04488) (1.225,331.20659) (1.275,271.50553) (1.725,611.7644) (1.775,562.44201) (1.825,568.47756) (1.875,589.80819) (1.925,558.90899) (1.975,590.47697) (2.025,593.09206) (2.075,583.29541) (2.125,543.68268) (2.175,563.67686) (2.225,595.48185) (2.275,533.56125) (2.725,583.26827) (2.775,538.30191) (2.825,569.10391) (2.875,527.07427) (2.925,537.85261) (2.975,546.49743) (3.025,528.84705) (3.075,553.30644) (3.125,535.43399) (3.175,544.39888) (3.225,517.15817) (3.275,526.14264) (3.725,549.99753) (3.775,500.2833) (3.825,496.66154) (3.875,457.58774) (3.925,503.51787) (3.975,506.32214) (4.025,493.34665) (4.075,473.43781) (4.125,432.23246) (4.175,486.57109) (4.225,290.58407) (4.275,9.36629) (4.725,458.95678) (4.775,444.32638) (4.825,439.92607) (4.875,436.43532) (4.925,419.0082) (4.975,449.30556) (5.025,445.55737) (5.075,429.54945) (5.125,406.64396) (5.175,425.80831) (5.225,413.72093) (5.275,446.77922) (5.725,371.51317) (5.775,359.54518) (5.825,352.64845) (5.875,395.36762) (5.925,350.18777) (5.975,353.88937) (6.025,370.82122) (6.075,360.39817) (6.125,348.76802) (6.175,348.91992) (6.225,356.68577) (6.275,341.05313) (6.725,201.58053) (6.775,335.34638) (6.825,302.60896) (6.875,303.53504) (6.925,287.31907) (6.975,287.62864) (7.025,299.73859) (7.075,286.46456) (7.125,282.3745) (7.175,296.69049) (7.225,271.58225) (7.275,313.50027) (7.725,0) (7.775,3.70101) (7.825,21.96833) (7.875,19.34601) (7.925,16.31272) (7.975,17.00493) (8.025,9.75488) (8.075,18.69411) (8.125,22.01666) (8.175,25.26713) (8.225,15.02581) (8.275,0)};
\draw[blue!70!black, line width=1.2pt] (axis cs:0.7,325.961695) -- (axis cs:1.3,325.961695);
\draw[blue!70!black, line width=1.2pt] (axis cs:1.7,575.886485) -- (axis cs:2.3,575.886485);
\draw[blue!70!black, line width=1.2pt] (axis cs:2.7,538.07726) -- (axis cs:3.3,538.07726);
\draw[blue!70!black, line width=1.2pt] (axis cs:3.7,489.95887) -- (axis cs:4.3,489.95887);
\draw[blue!70!black, line width=1.2pt] (axis cs:4.7,438.180695) -- (axis cs:5.3,438.180695);
\draw[blue!70!black, line width=1.2pt] (axis cs:5.7,355.28757) -- (axis cs:6.3,355.28757);
\draw[blue!70!black, line width=1.2pt] (axis cs:6.7,292.159565) -- (axis cs:7.3,292.159565);
\draw[blue!70!black, line width=1.2pt] (axis cs:7.7,16.658825) -- (axis cs:8.3,16.658825);
\nextgroupplot[title={(b) Model-based decision scenarios},
ymin=0.5, ymax=4.5, ytick={1,2,3,4},
yticklabels={Direct product,Direct residual,Feed-norm. product,Feed-norm. residual},
yticklabel style={font=\fontsize{7}{8}\selectfont},
grid=major, grid style={gray!20},
legend style={at={(0.5,-0.32)}, anchor=north, legend columns=1, draw=none, font=\scriptsize}]
\addplot[only marks, mark=square*, mark size=3.3pt, mark options={fill=blue!18, draw=blue!65!black}] coordinates {(2,1) (5,2) (6,2) (7,2) (8,2) (2,3) (3,3) (4,3) (5,3) (7,3) (3,4) (4,4) (6,4) (7,4) (8,4)};
\addlegendentry{Mg--Ca frontier}
\addplot[only marks, mark=diamond*, mark size=3.3pt, mark options={fill=orange!80, draw=orange!75!black}] coordinates {(2,1) (8,2) (7,3) (8,4)};
\addlegendentry{Mg-only leader}
\end{groupplot}
\end{tikzpicture}
\caption{Produced-water replication and interpretation sensitivity. Panel (a) retains all 12 recorded Mg values per row (gray) and shows their medians (blue). Panel (b) gives Model-based Mg--Ca frontiers; diamonds mark the Mg-only leader. Rows jointly vary NaOH and produced-water volumes, and signal units and measured phase remain unconfirmed.}
\label{fig:produced-water-interpretation}
\end{figure}

\textsc{Derived}. The direct (un-normalized) Mg medians decrease in the order B, C, D, E, F, A, G, H (Fig.~\ref{fig:produced-water-interpretation}). \textsc{Model-based}. If signals are condition-comparable, omitting any one well leaves this order unchanged (minimum Spearman $\rho_s=1$, maximum rank shift 0). Under the additional common-feed and linear-response assumptions required for feed-fraction normalization, the product-phase leader changes from B to G; if the signals are instead read as residual-phase measurements, both the direct and the feed-normalized readings select row H. The Mg--Ca frontiers also change. Replication therefore does not identify a preferred row without knowing the measured phase and dilution corrections.

The matched ratios motivate a scale-transfer question; they do not show that the plate selected the bench condition or establish plant-scale transfer.

\subsection{Other records}

\textsc{Observed}. The saved SmCo BO state specifies a Mat\'ern-2.5 model, fixed-low noise, UCB with $\beta=0.2$, and a 96-point terminal batch. Its optimization history is empty, so it documents the configuration but lacks complete selection provenance. The deposited TEA records identify possible products and missing inputs but do not define a numerical downstream loss or recorded feedback rule for selecting experiments. The supplementary information reports a UV--Vis versus ICP-MS slope of 0.97. Without paired values, residual calibration, turnaround, and cost, UV--Vis is a candidate alternative assay rather than a calibrated substitute.

\section{Decision-focused formulation and algorithmic direction}

\subsection{Downstream loss}

The SmCo example links a recorded condition to loss through nominal recovery and a grade requirement. We now extend that decision structure to uncertain technical responses, process costs, and the intended deployment scale.

Let $r\in\mathcal R$ denote a recovery route, $x\in\mathcal X_r$ an operating condition, and $s\in\mathcal S$ an experimental scale or assay fidelity (how faithfully the assay measures the quantity of interest). For action $a=(r,x,s)$,
\[
Y_a\sim p(\cdot\mid a,\theta),
\]
where $\theta$ contains uncertain technical quantities such as response surfaces, recovery, impurity behavior, measurement noise, and scale discrepancy. Let $\phi$ contain exogenous economic quantities such as product value, reagent price, disposal cost, and throughput value. A deployment decision $d\in\mathcal D$ is evaluated at target scale $s_\star$ by
\[
L_{s_\star}(d,\theta,\phi)
=C_{\mathrm{process}}(d,\theta,\phi;s_\star)
+C_{\mathrm{penalty}}(d,\theta,\phi;s_\star).
\]
Applying this loss to a selected campaign requires defining non-overlapping terms.

Let $b_t(\theta)$ be the technical posterior after $t$ batches and $q(\phi)$ an economic scenario distribution. The notation treats $\phi$ as exogenous to $\theta$; a joint distribution can replace it if dependence matters. Current Bayes risk is
\[
R_{s_\star}(b_t,q)
=\min_{d\in\mathcal D}
\mathbb E_{\theta\sim b_t,\,\phi\sim q}
\big[L_{s_\star}(d,\theta,\phi)\big].
\]
Under this factorized model, laboratory observations update $b_t$ but not $q$ unless an action measures an economic quantity. Under a joint model, they may also update beliefs about correlated components of $\phi$.

\subsection{Value of a batch}

For feasible batch $B$ with future outcome $Y_B$, define
\[
\Delta(B\mid b_t,q,s_\star)
=R_{s_\star}(b_t,q)
-\mathbb E_{Y_B\mid b_t}
\left[R_{s_\star}(b_{t+1}^{B,Y_B},q)\right].
\]
Under exact Bayesian updating and optimization, $\Delta(B)\ge 0$ because the decision maker can ignore an observation. This normative value-of-information objective \citep{lindley1956measure,frazier2008knowledge} still requires a tractable acquisition rule for selecting 96-well batches.

If every plate has capacity $Q$ and the same cost, the primary selection problem is
\[
B_t^*\in\arg\max_{B\in\mathcal B_Q}\Delta(B\mid b_t,q,s_\star),
\]
where $\mathcal B_Q$ contains physical, timing, control, and replication constraints. Subtracting a common plate cost does not change the ranking. Experimental cost matters when comparing unequal fidelities, variable batch sizes, turnaround, or stopping. It can then enter as a constraint $c_{\mathrm{exp}}(B)\le C_t$ or as $\Delta(B)-c_{\mathrm{exp}}(B)$ after conversion to common loss units.

\subsection{Candidate method family}

The exact method is open. Three paths are plausible:

\begin{enumerate}[leftmargin=1.5em,itemsep=2pt]
\item \textbf{Downstream-aware two-stage policy:} retain discrimination followed by within-route optimization, but define both around downstream loss.
\item \textbf{Joint decision-value policy:} optimize approximate $\Delta(B)$ directly over route--condition actions.
\item \textbf{Hybrid structured policy:} use \eC{}-style route discrimination and route-specific GP criteria to generate candidates, then allocate the finite batch by approximate downstream value.
\end{enumerate}

The hybrid is a plausible candidate because it retains the companion preprint's useful structure without adding an \eC{} impurity score directly to a GP-UCB score. It would fit route-specific posteriors, group plausible scenarios sampled from those posteriors by their preferred route, generate conditions near promising or decision-sensitive regions, and allocate wells across routes, conditions, replicates, and controls by approximate marginal $\Delta$. Posterior representation, batching, and any theoretical result should follow from the selected implementation.

\section{Retrospective analyses and process decision model}

\subsection{Valid retrospective evaluation}

An observation-based policy may select only recorded CICERO conditions. Off-grid outcomes require a response model and must be labeled \textsc{Model-based}.

The produced-water plate has 12 recorded wells at each of eight conditions. Completed leave-one-out and prespecified outlier checks are confined to those conditions; centered column ranks are descriptive and do not identify causal position effects. SmCo and NdFeB generally have one outcome per condition. They support benchmarks that reveal entries from the completed grid, not repeated noisy trajectories or claims about the policy that originally generated the grid.

\subsection{Conditional NdFeB finite-pool benchmark}

Each policy selects recorded conditions and then sees their stored outcomes. We measure how many revealed wells it requires to find the exact maximum among the fixed recorded $E$ values, at H10. The route map and enrichment definition follow the assumptions in Section~\ref{sec:ndfeb-records}. GP predictions guide selection; stored outcomes determine search performance. This differs from the quadratic-ridge interpolation diagnostic in that section. Within the recorded pool, an explicit downstream loss and sufficient decision inputs could support a different ranking.

The two-stage reconstruction replaces the companion method's \eC{} diagnostic stage with an initial batch split equally between routes. Using separate GPs fitted to those results, it commits to the route with the highest posterior mean log enrichment at any of its recorded conditions, then uses GP-UCB within that route. Routewise GP-UCB lets separate route models compete without committing to one route; mixed-route BO uses one joint route--condition model. Equal route splitting uses GP-UCB adaptively within each route.

The space-filling baseline spreads selections across the recorded grid without using outcomes. Each next point is as far as possible from its nearest previously selected point in normalized route--condition coordinates.

\textsc{Model-based}. Under fitted GP hyperparameters, the two-stage reconstruction, routewise GP-UCB, and equal route split reveal H10 by 16 wells. Mixed-route BO does so by 24 and deterministic space filling by 48. This comparison supports faster discovery by adaptive search on this pool. Random sampling falls short of the pool-optimum enrichment of 508.97 by 21.36 on average (grid simple regret) after 48 wells. Figure~\ref{fig:ndfeb-finite-pool} shows the search trajectories; per-budget metrics are tabulated in Appendix~\ref{app:finite-pool}.

The first three adaptive policies tie on best-revealed enrichment throughout the benchmark. Their common initial batch reveals H12 ($E=493.11$), near the pool maximum 508.97. Space filling also reveals H12 in its initial batch; different initial designs prevent isolating the effect of adaptation alone.

The three adaptive policies allocate differently: by 48 wells they assign 44, 43, and 24 wells, respectively, to oxalate. Once H10 is revealed, best-revealed enrichment is saturated and cannot value those differences. The terminal-only symbolic shortfall $\max(0,T-E)$, for a fixed enrichment target $T$, ties all conditions with $E\geq T$ and does not drive acquisition; the benchmark therefore does not test downstream-aware selection.

\textsc{Model-based}. Plug-in latent intervals describe uncertainty in the underlying log-enrichment response with fitted GP hyperparameters held fixed. At 16 wells, nominal 90\% plug-in latent intervals cover 42.5\% of the 80 hidden recorded conditions under both the two-stage and routewise models (identical coverage), so the trajectories do not establish calibrated uncertainty reduction.

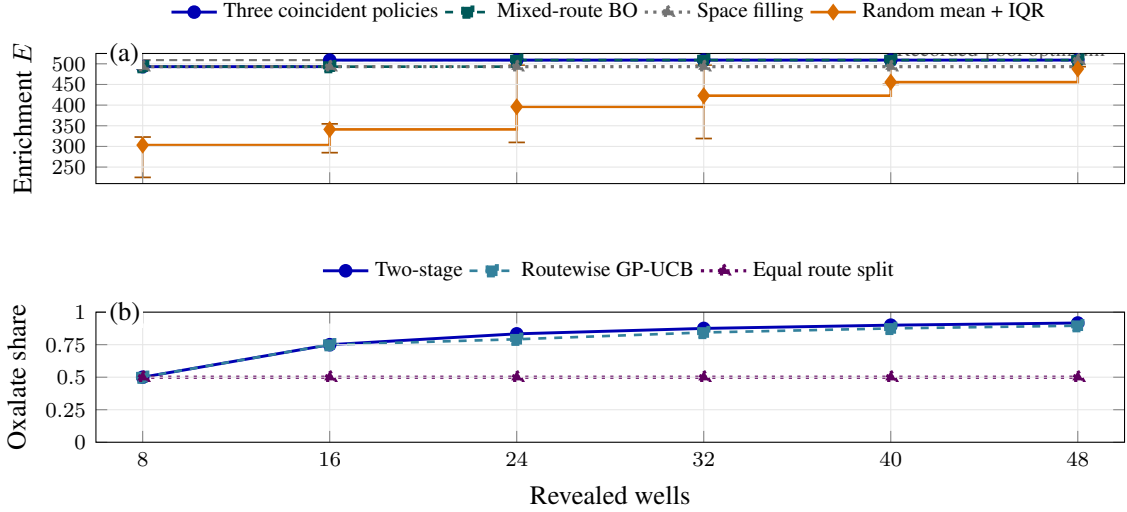
\begin{figure}[ht]
\centering
\begin{tikzpicture}
\begin{groupplot}[
group style={group size=1 by 2, vertical sep=1.7cm},
width=0.92\textwidth, height=0.20\textwidth,
xmin=6, xmax=50, xtick={8,16,24,32,40,48},
tick label style={font=\scriptsize}, label style={font=\small},
title style={at={(0.01,0.98)}, anchor=north west, font=\small,
fill=white, inner sep=1pt}, grid=major, grid style={gray!20}, axis on top
]
\nextgroupplot[title={(a)},
ymin=210, ymax=525, ytick={250,300,350,400,450,500},
xticklabels=\empty,
ylabel={Enrichment $E$},
legend style={at={(0.5,1.16)}, anchor=south,
legend columns=4, draw=none, font=\scriptsize}]
\draw[orange!65!black, line width=0.65pt] (axis cs:8,225.000078) -- (axis cs:8,322.714061);
\draw[orange!65!black, line width=0.65pt] (axis cs:7.65,225.000078) -- (axis cs:8.35,225.000078);
\draw[orange!65!black, line width=0.65pt] (axis cs:7.65,322.714061) -- (axis cs:8.35,322.714061);
\draw[orange!65!black, line width=0.65pt] (axis cs:16,284.8397) -- (axis cs:16,354.604786);
\draw[orange!65!black, line width=0.65pt] (axis cs:15.65,284.8397) -- (axis cs:16.35,284.8397);
\draw[orange!65!black, line width=0.65pt] (axis cs:15.65,354.604786) -- (axis cs:16.35,354.604786);
\draw[orange!65!black, line width=0.65pt] (axis cs:24,309.695735) -- (axis cs:24,497.07493);
\draw[orange!65!black, line width=0.65pt] (axis cs:23.65,309.695735) -- (axis cs:24.35,309.695735);
\draw[orange!65!black, line width=0.65pt] (axis cs:23.65,497.07493) -- (axis cs:24.35,497.07493);
\draw[orange!65!black, line width=0.65pt] (axis cs:32,319.14451) -- (axis cs:32,497.07493);
\draw[orange!65!black, line width=0.65pt] (axis cs:31.65,319.14451) -- (axis cs:32.35,319.14451);
\draw[orange!65!black, line width=0.65pt] (axis cs:31.65,497.07493) -- (axis cs:32.35,497.07493);
\draw[orange!65!black, line width=0.65pt] (axis cs:40,450.27696) -- (axis cs:40,508.966758);
\draw[orange!65!black, line width=0.65pt] (axis cs:39.65,450.27696) -- (axis cs:40.35,450.27696);
\draw[orange!65!black, line width=0.65pt] (axis cs:39.65,508.966758) -- (axis cs:40.35,508.966758);
\draw[orange!65!black, line width=0.65pt] (axis cs:48,493.110987) -- (axis cs:48,508.966758);
\draw[orange!65!black, line width=0.65pt] (axis cs:47.65,493.110987) -- (axis cs:48.35,493.110987);
\draw[orange!65!black, line width=0.65pt] (axis cs:47.65,508.966758) -- (axis cs:48.35,508.966758);
\addplot[black!55, densely dashed, line width=0.8pt, forget plot] coordinates {(8,508.966758) (16,508.966758) (24,508.966758) (32,508.966758) (40,508.966758) (48,508.966758)};
\node[anchor=south east, font=\scriptsize, text=black!65, fill=white, inner sep=1pt] at (axis cs:49.3,500) {Recorded-pool optimum};
\addplot[const plot, blue!70!black, line width=1.1pt, mark=*, mark size=2pt] coordinates {(8,493.110987) (16,508.966758) (24,508.966758) (32,508.966758) (40,508.966758) (48,508.966758)};
\addlegendentry{Three coincident policies}
\addplot[const plot, teal!70!black, dashed, line width=1pt, mark=square*, mark size=2pt] coordinates {(8,493.110987) (16,493.110987) (24,508.966758) (32,508.966758) (40,508.966758) (48,508.966758)};
\addlegendentry{Mixed-route BO}
\addplot[const plot, black!55, dotted, line width=1.2pt, mark=triangle*, mark size=2.4pt] coordinates {(8,493.110987) (16,493.110987) (24,493.110987) (32,493.110987) (40,493.110987) (48,508.966758)};
\addlegendentry{Space filling}
\addplot[const plot, orange!85!black, line width=1pt, mark=diamond*, mark size=2.4pt] coordinates {(8,303.144705) (16,341.201453) (24,395.967717) (32,422.837208) (40,455.620163) (48,487.605705)};
\addlegendentry{Random mean + IQR}
\nextgroupplot[title={(b)},
ymin=0, ymax=1, ytick={0,0.25,0.5,0.75,1},
xlabel={Revealed wells}, ylabel={Oxalate share},
legend style={at={(0.5,1.16)}, anchor=south,
legend columns=3, draw=none, font=\scriptsize}]
\addplot[blue!70!black, line width=1.1pt, mark=*, mark size=2pt] coordinates {(8,0.5) (16,0.75) (24,0.833333333) (32,0.875) (40,0.9) (48,0.916666667)};
\addlegendentry{Two-stage}
\addplot[cyan!55!black, dashed, line width=1pt, mark=square*, mark size=2pt] coordinates {(8,0.5) (16,0.75) (24,0.791666667) (32,0.84375) (40,0.875) (48,0.895833333)};
\addlegendentry{Routewise GP-UCB}
\addplot[violet!75!black, dotted, line width=1.2pt, mark=triangle*, mark size=2.4pt] coordinates {(8,0.5) (16,0.5) (24,0.5) (32,0.5) (40,0.5) (48,0.5)};
\addlegendentry{Equal route split}
\end{groupplot}
\end{tikzpicture}
\caption{Model-based reveal-only search on the conditional NdFeB Round~1 pool under fitted GP hyperparameters. Panel (a) shows the best recorded enrichment revealed; random points are means with interquartile-range bars over 20 seeds, and space filling is an outcome-blind maximin baseline. Panel (b) shows the cumulative oxalate share for the three policies with coincident incumbent trajectories.}
\label{fig:ndfeb-finite-pool}
\end{figure}

\subsection{Decision inputs and evidence requirements}

The downstream loss should contain only quantities that can affect the deployment choice, value of information, or stopping. Laboratory-resolvable quantities include the executed protocol, measured phase, units and dilution, recovery, grade or selectivity, replicate variation, and condition--response relationships.

\textsc{PNNL-confirmed}. CICERO's precipitation operations were chosen for transferability to conventional tank reactors. Scale-transfer quantities include mixing, mass transfer, residence time, reactor geometry, solid--liquid handling, throughput, and process robustness. Validating pilot or plant performance requires experiments at the corresponding scales; extrapolations remain \textsc{Model-based}. Scale affects both technical response and the consequence of deployment; these effects should not be collapsed into one multiplier.

Economic and decision inputs include product specifications; the marginal value of higher purity; reagent, waste, and refining costs; throughput requirements; customer requirements; and market conditions. Because these inputs may be incomplete or time-varying, the analysis should use credible ranges, scenarios, qualitative orderings, or symbolic decision boundaries.

Laboratory observations inform technical responses, and paired multi-scale experiments inform scale transfer. Under the factorized model, economic beliefs remain fixed unless an action measures an economic quantity. The analysis may identify a robust, viable process choice or a missing input that could change that choice.

\section{Exploratory synthetic design probe}

\textsc{Model-based}. We compared implementations of the candidate architectures under a common normalized downstream loss. The synthetic two-stage policy starts with a balanced batch, commits to the route whose best condition has the lowest estimated expected downstream loss, then selects experiments by approximate decision value within that route. The joint decision-value policy selects across routes and conditions. The structured hybrid proxy first filters for promising or uncertain candidates, then applies the joint policy's decision-value rule. This proxy is not the proposed \eC{}/GP-UCB hybrid.

The full 32-cell factorial spans route gap, batch capacity, scale discrepancy, loss shape, and grade-specification uncertainty. Each cell used the same 100 world seeds, a resource budget of 16, and laboratory and bench resource costs of one and two. Particles are posterior samples of model coefficients: we used 256 acquisition particles for experiment selection and 4{,}096 terminal particles for the final process decision.

Excess loss is the downstream loss of the chosen action minus that of a synthetic oracle. On this normalized scale, not deploying has loss one. Table~\ref{tab:synthetic-contrasts} reports paired excess-loss differences averaged across cells; world-seed Monte Carlo standard errors use 50 balanced adjacent-seed blocks. Mean excess losses are 0.041 for the structured hybrid proxy, 0.045 for synthetic two-stage, and 0.046 for the joint decision-value policy. The largest contrast is roughly a tenth of the policies' mean excess loss.

\begin{table}[ht]
\centering
\caption{Exploratory synthetic policy contrasts.}
\label{tab:synthetic-contrasts}
\begin{tabular}{lrr}
\toprule
Paired contrast & Mean difference & World-MC SE \\
\midrule
Structured hybrid proxy $-$ Two-stage & -0.0039 & 0.0051 \\
Structured hybrid proxy $-$ Joint decision-value & -0.0043 & 0.0019 \\
Joint decision-value $-$ Two-stage & 0.0004 & 0.0047 \\
\bottomrule
\end{tabular}
\par{\footnotesize Negative values favor the first-named policy. Standard errors are conditional on the configured synthetic model and acquisition approximation.\par}

\end{table}

The structured hybrid proxy has lower estimated loss than the direct joint decision-value policy; both contrasts involving synthetic two-stage are small relative to their world-seed Monte Carlo standard errors. Across 64, 128, and 256 acquisition particles, the structured-minus-two-stage and structured-minus-joint estimates remain negative, while joint-minus-two-stage remains near zero. Across particle counts the aggregate pattern is stable; individual acquisition paths are not. These simulations contain no empirical target-scale inputs. The next step is to select an acquisition method for the prospective comparison.

\section{Proposed prospective campaign}

A prospective campaign should use a feedstock with at least two credible routes and continuous conditions within each route. It should be small enough for two or three adaptive batches but large enough that exhaustive testing is impractical.

Before the first batch, a prospective campaign should pre-register the route set, condition bounds, measured outputs, downstream loss or economic scenarios, candidate pool, physical constraints, common budget, controls, stopping rule, and evaluation metrics. The pre-registration should also state that an unchanged preferred decision is a valid result.

The first batch should cover routes, feasibility, and the downstream decision boundary. The second should compare the incumbent proposal with the decision-focused allocation using the same information. Where practical, both recommendations should share a plate to reduce batch confounding. Bench or pilot confirmation should carry forward the selected condition and at least one credible alternative.

For each round, retain the candidate set, exclusions, code and configuration versions, random seeds, model state, acquisition values, proposed and executed batches, manual changes, timestamps, resource use, raw measurements, dilution and QC metadata, and stopping decision. Use versioned pipelines and interoperable schemas to convert these records into decision-ready variables.

\section{Related work, limitations, and open questions}

The proposed objective draws on Bayesian experimental design and knowledge-gradient methods, which value observations by their expected effect on later decisions \citep{lindley1956measure,frazier2008knowledge}. The companion preprint supplies a two-stage structure: equivalence-class discrimination followed by within-route GP optimization \citep{ray2026coactive}. \eC{} addresses Bayesian decision-region identification \citep{golovin2010near}; GP-UCB and related methods address expensive black-box optimization \citep{srinivas2010gaussian,frazier2018tutorial}. Batch and cost-aware BO address parallel evaluations and differences in experimental cost \citep{desautels2014parallelizing,astudillo2021multistep}. A detailed comparison with batch knowledge gradient, terminal Bayesian optimization, decision-focused design, technoeconomic-aware optimization, and multi-fidelity design will depend on the acquisition method selected.

The CICERO archive supports conditional comparisons among recorded conditions. It cannot directly validate policies that select unrecorded conditions: better metadata would improve reconstruction but would not supply the missing outcomes. Most conditions lack independent replicates, limiting estimates of repeatability. Policy trajectories are unavailable, and changes between rounds may combine chemistry and protocol effects, limiting reconstruction of the original campaign. The archive also lacks a validated downstream loss and sufficient data to establish plant-scale transfer.

Several measurement and provenance details remain unresolved in the available files and publications. Author confirmation or additional records could resolve them; they are likely known to the experimental team, but confirmation was limited by the deadline for submitting the current manuscript.

For SmCo, interpreting nominal yield requires confirming whether the 3928.381~mg/L feed concentration applies to both rounds and how nominal yields above 100\% should be treated. Interpreting the repeated Co value $4.8983050847\times10^{-10}$~mol requires confirming whether it is an analytical floor. Reconstructing the Round~2 selection also requires the unavailable BO recommendations and their mapping to executed wells.

For NdFeB, interpreting the route comparison requires the executed plate maps, reagent concentrations, measured phases, dilution treatment, and intended separation-factor definition. Calculating recovery across rounds additionally requires the enriched-stock preparation and recovery denominator.

For produced water, condition rankings depend on the executed map and stock concentration, signal units, dilution treatment, and measured phase. Assessing replication requires confirming whether columns are replicates and clarifying any outlier treatment. Linking plate row~B to the bench experiment requires confirming whether it informed that experiment and whether the feed was materially comparable.

Choosing a process also requires decision inputs: the Sm product-grade threshold, rework cost, and value of recovered Sm. The NdFeB enrichment score alone omits recovery, reagent demand, throughput, waste, and downstream purification. These inputs are distinct from the measurement and provenance gaps, which motivate the data infrastructure and logging requirements specified above.

\section{Conclusion}

In the conditional \textsc{Model-based} NdFeB benchmark, adaptive policies reach the recorded enrichment maximum by 16--24 wells, compared with 48 for space filling. Our two-stage reconstruction shares the earliest discovery budget of 16 wells with two adaptive alternatives.

Conditional retrospective analyses show a tradeoff between purity and nominal yield in SmCo, different enrichment profiles across NdFeB routes, and produced-water condition rankings that depend on how the measurements are interpreted. Selecting an economically preferred process requires an explicit downstream loss and the inputs needed to evaluate it. We propose using expected reduction in downstream Bayes risk to select experimental batches.

The prospective comparison would test batch selection under a common downstream loss and logging standard. Preparing it requires selecting an acquisition method, resolving the archive's measurement and provenance gaps, and defining the deployment decision, relevant outputs, and credible economic ranges.

\section*{Acknowledgments}

We thank the PNNL CICERO team, whose published study and deposited records made these analyses possible. We are also grateful to Chinmayee Subban, Grant Johnson, Maxim Ziatdinov, and Andrew Ritchhart of Pacific Northwest National Laboratory. E.N. was supported by the Foundational Autonomy Investment (FAI) under the Laboratory Directed Research and Development (LDRD) Program at Pacific Northwest National Laboratory. PNNL is a multi-program national laboratory operated for the U.S. Department of Energy by Battelle Memorial Institute under Contract No.~DE-AC05-76RL01830.

\appendix
\section{Finite-pool policy table}\label{app:finite-pool}

\begin{table}[ht]
\centering
\caption{Model-based reveal-only NdFeB finite-pool grid simple regret by well budget. Random-design entries average 20 seeded runs; other policies are deterministic. Values measure enrichment shortfall from the recorded-pool optimum.}
\label{tab:ndfeb-finite-pool}
\scriptsize
\begin{tabular}{lrrrrrrr}
\toprule
Policy & Runs & 8 & 16 & 24 & 32 & 40 & 48 wells \\
\midrule
Two-stage reconstruction & 1 & 15.86 & 0.00 & 0.00 & 0.00 & 0.00 & 0.00 \\
Routewise GP-UCB & 1 & 15.86 & 0.00 & 0.00 & 0.00 & 0.00 & 0.00 \\
Equal route split & 1 & 15.86 & 0.00 & 0.00 & 0.00 & 0.00 & 0.00 \\
Mixed-route BO & 1 & 15.86 & 15.86 & 0.00 & 0.00 & 0.00 & 0.00 \\
Random & 20 & 205.82 & 167.77 & 113.00 & 86.13 & 53.35 & 21.36 \\
Space filling & 1 & 15.86 & 15.86 & 15.86 & 15.86 & 15.86 & 0.00 \\
Oracle & 1 & 0.00 & 0.00 & 0.00 & 0.00 & 0.00 & 0.00 \\
\bottomrule
\end{tabular}
\end{table}

\noindent Grid simple regret uses the best revealed condition. Symbolic downstream metrics use each policy's permitted terminal action; the two-stage terminal action remains within its committed route.

\section{Robustness and descriptive analyses}

\paragraph{SmCo.} \textsc{Model-based}. A common positive feed-denominator multiplier leaves raw Pareto membership unchanged. Capping or excluding nominal yields above 100\% changes some frontier membership but not the five-point Round~2 tradeoff at or below 100\%; the main analysis retains raw nominal values.

\paragraph{Produced water.} \textsc{Model-based}. The Mg-only leaders and Mg--Ca frontiers are unchanged under the retain-all, Tukey 1.5-IQR, and modified-$z$ 3.5 rules. These rules are sensitivity checks; the main analysis retains all measurements.

\paragraph{NdFeB Round 2.} \textsc{Derived}. Within-plate row and column profiles retain all 96 recorded values, including finite negative signals. No cross-round recovery or material balance is calculated because the enriched-stock denominator is unavailable.

\paragraph{Finite-pool sensitivity.} \textsc{Model-based}. Across the fitted and four perturbed GP specifications, the first grid-optimum discovery occurs at 16--24 wells for the two-stage and routewise policies, 16--40 for equal route splitting, and 24--40 for mixed-route BO. These scenarios probe model sensitivity; they do not provide uncertainty intervals.

\paragraph{Synthetic particle sensitivity.} \textsc{Model-based}.
\begin{center}
\begin{tabular}{lrrr}
\toprule
Paired contrast & 64 particles & 128 particles & 256 particles \\
\midrule
Structured hybrid proxy $-$ Two-stage & -0.0057 & -0.0013 & -0.0039 \\
Structured hybrid proxy $-$ Joint decision-value & -0.0052 & -0.0035 & -0.0043 \\
Joint decision-value $-$ Two-stage & -0.0005 & 0.0022 & 0.0004 \\
\bottomrule
\end{tabular}
\par{\footnotesize Point estimates only; negative values favor the first-named policy.\par}

\end{center}

{\small
\bibliographystyle{plainnat}
\bibliography{references}

\begin{thebibliography}{9}
\providecommand{\natexlab}[1]{#1}
\providecommand{\url}[1]{\texttt{#1}}
\expandafter\ifx\csname urlstyle\endcsname\relax
  \providecommand{\doi}[1]{doi: #1}\else
  \providecommand{\doi}{doi: \begingroup \urlstyle{rm}\Url}\fi

\bibitem[Astudillo et~al.(2021)Astudillo, Jiang, Balandat, Bakshy, and
  Frazier]{astudillo2021multistep}
Ra{\'u}l Astudillo, Daniel~R. Jiang, Maximilian Balandat, Eytan Bakshy, and
  Peter~I. Frazier.
\newblock Multi-step budgeted bayesian optimization with unknown evaluation
  costs.
\newblock In \emph{Advances in Neural Information Processing Systems}, 2021.

\bibitem[Desautels et~al.(2014)Desautels, Krause, and
  Burdick]{desautels2014parallelizing}
Thomas Desautels, Andreas Krause, and Joel~W. Burdick.
\newblock Parallelizing exploration-exploitation tradeoffs in gaussian process
  bandit optimization.
\newblock \emph{Journal of Machine Learning Research}, 15:\penalty0 3873--3923,
  2014.

\bibitem[Frazier(2018)]{frazier2018tutorial}
Peter~I. Frazier.
\newblock A tutorial on bayesian optimization.
\newblock \emph{arXiv preprint arXiv:1807.02811}, 2018.

\bibitem[Frazier et~al.(2008)Frazier, Powell, and
  Dayanik]{frazier2008knowledge}
Peter~I. Frazier, Warren~B. Powell, and Savas Dayanik.
\newblock A knowledge-gradient policy for sequential information collection.
\newblock \emph{SIAM Journal on Control and Optimization}, 47\penalty0
  (5):\penalty0 2410--2439, 2008.

\bibitem[Golovin et~al.(2010)Golovin, Krause, and Ray]{golovin2010near}
Daniel Golovin, Andreas Krause, and Debajyoti Ray.
\newblock Near-optimal bayesian active learning with noisy observations.
\newblock In \emph{Advances in Neural Information Processing Systems}, 2010.

\bibitem[Lindley(1956)]{lindley1956measure}
Dennis~V. Lindley.
\newblock On a measure of the information provided by an experiment.
\newblock \emph{The Annals of Mathematical Statistics}, 27\penalty0
  (4):\penalty0 986--1005, 1956.

\bibitem[Ray and Srinivas(2026)]{ray2026coactive}
Debajyoti Ray and Niranjan Srinivas.
\newblock Cost-aware recovery-pathway identification and bayesian optimization
  for autonomous materials discovery.
\newblock \emph{arXiv preprint arXiv:2607.23896}, 2026.

\bibitem[Ritchhart et~al.(2026)Ritchhart, Allec, Butreddy, Kulesa, Wang,
  Nguyen, Ziatdinov, and Nakouzi]{ritchhart2026materials}
Andrew Ritchhart, Sarah~I. Allec, Pravalika Butreddy, Krista Kulesa, Qingpu
  Wang, Dan~Thien Nguyen, Maxim Ziatdinov, and Elias Nakouzi.
\newblock Agentic workflow enables the recovery of critical materials from
  complex feedstocks via selective precipitation.
\newblock \emph{Materials Horizons}, 2026.
\newblock \doi{10.1039/D6MH00475J}.
\newblock Advance Article; first published 21 May 2026.

\bibitem[Srinivas et~al.(2010)Srinivas, Krause, Kakade, and
  Seeger]{srinivas2010gaussian}
Niranjan Srinivas, Andreas Krause, Sham~M. Kakade, and Matthias Seeger.
\newblock Gaussian process optimization in the bandit setting: No regret and
  experimental design.
\newblock In \emph{International Conference on Machine Learning}, 2010.

\end{thebibliography}
}

\end{document}